\documentclass[conference,a4paper]{IEEEtran}
\IEEEoverridecommandlockouts
\usepackage{cite}
\usepackage{amsmath,amssymb,amsfonts}
\usepackage{algorithmic}
\usepackage{graphicx}
\usepackage{textcomp}
\usepackage{xcolor}
\usepackage{tabularx}
\usepackage{multirow}
\RequirePackage{float}
\usepackage[colorinlistoftodos]{todonotes}
\newlength{\preferredwidth}
\newcommand{\x}{{\mathbf{x}}}

\newcommand{\tb}{{\mathbf{{t}}}}
\newcommand{\ub}{{\mathbf{{u}}}}
\newcommand{\vb}{{\mathbf{{v}}}}
\newcommand{\R}{{\mathbb{R}}}
\usepackage{adjustbox}
\usepackage[T1]{fontenc}
\usepackage[utf8]{inputenc}
\usepackage[ colorlinks=true,pdfborder={0 0 0},linkcolor=blue]{hyperref}
\RequirePackage{enumitem}
\RequirePackage{booktabs} 
\usepackage{threeparttable}
\usepackage{microtype} 

\begin{document}
\title{Towards a Real-Time Facial Analysis System}
\author{
Bishwo Adhikari$^1$ \quad Xingyang Ni$^1$ \quad Esa Rahtu$^1$ \quad Heikki Huttunen$^2$\\
$^1$Tampere University, Finland \quad $^2$Visy Oy, Finland\\
{\tt\small \{bishwo.adhikari, xingyang.ni, esa.rahtu\}@tuni.fi \quad heikki.huttunen@visy.fi}
}

\maketitle

\begin{abstract}
Facial analysis is an active research area in computer vision, with many practical applications.
Most of the existing studies focus on addressing one specific task and maximizing its performance.
For a complete facial analysis system, one needs to solve these tasks efficiently to ensure a smooth experience.
In this work, we present a system-level design of a real-time facial analysis system.
With a collection of deep neural networks for object detection, classification, and regression, the system recognizes age, gender, facial expression, and facial similarity for each person that appears in the camera view.
We investigate the parallelization and interplay of individual tasks.
Results on common off-the-shelf architecture show that the system's accuracy is comparable to the state-of-the-art methods, and the recognition speed satisfies real-time requirements.
Moreover, we propose a multitask network for jointly predicting the first three attributes, i.e., age, gender, and facial expression.
Source code and trained models are available at \url{https://github.com/mahehu/TUT-live-age-estimator}.
\end{abstract}

\begin{IEEEkeywords}
face detection, face recognition, facial similarity, real-time system
\end{IEEEkeywords}

\section{Introduction}
\label{intro}

Human facial analysis is one of most widely studied areas in computer vision, including topics such as face verification~\cite{DeepFace_face_verification, face_verification_2009}, head pose estimation~\cite{head_pose_survey, head_pose_cvpr_2011}, facial expression recognition~\cite{ facial_expression_survey_2009, expression_via_deep__belief_network} and age estimation~\cite{age_estimation_dl_2016}.
While computer programs have traditionally been unable to analyze facial images, humans are very good at spotting even the smallest differences. With the surge of deep learning techniques, algorithms have surpassed human accuracy in most of the above tasks.

In the field of facial image analysis, the majority of works focus on improving the accuracy of a specific task.
Less attention is paid to investigate the computational complexity~\cite{MobileNet, MobileNet_V2, efficientNets_2019}, in particular at the system level, where the architect needs to pay attention to the functionality of the entire system as well as that of the individual components; simultaneously optimizing for prediction accuracy, inference speed, memory footprint, parallelization as well as user experience (\textit{i.e.}, the system should at least \textit{appear} smooth although some components might operate at below real-time speed).

\begin{figure}[t]
\centering
\includegraphics[width=0.46\textwidth]{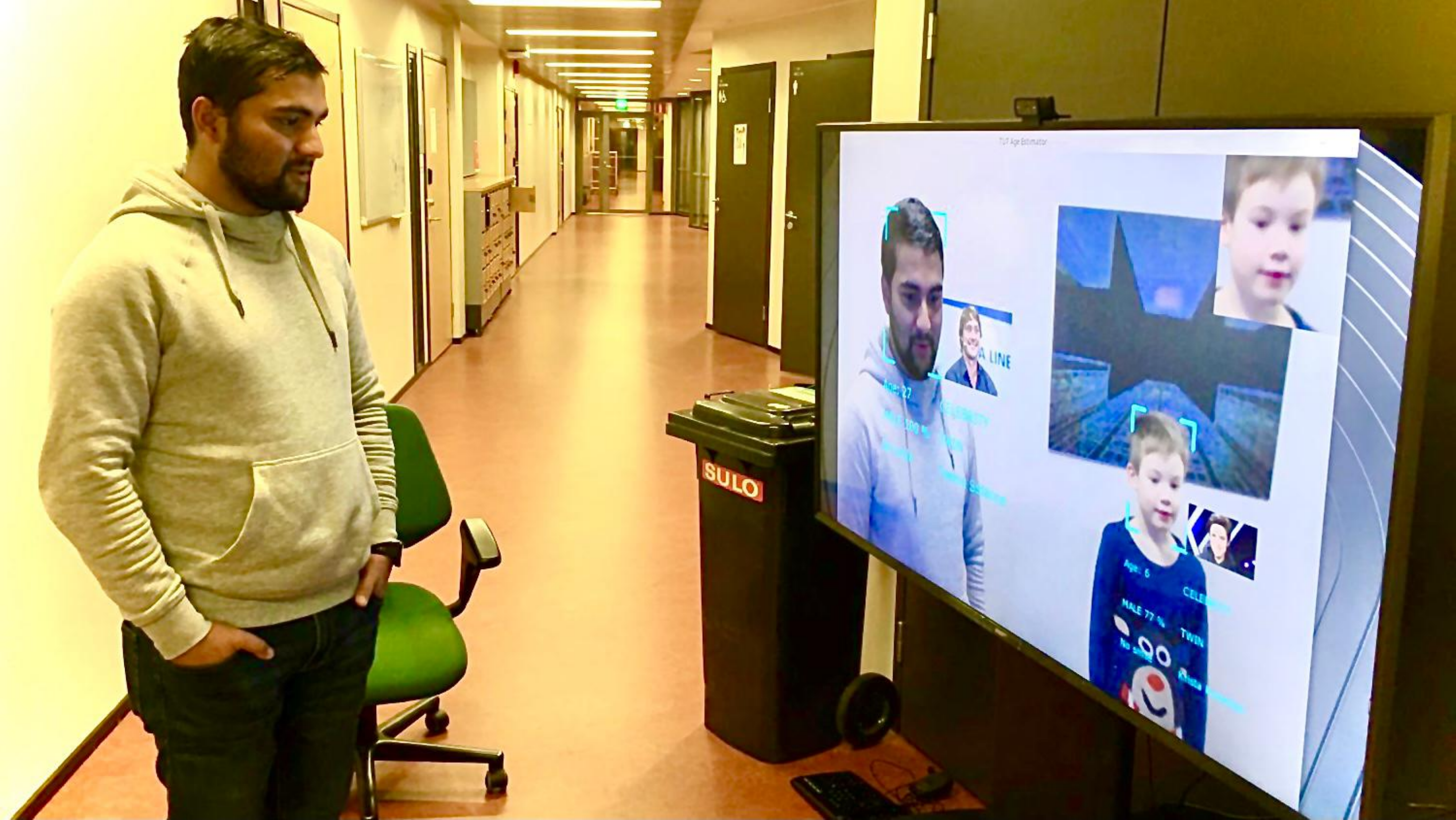}
\caption{
Our real-time facial recognition system in action.
It detects human faces on a frame captured by a webcam, recognizes age, gender, and emotion in real-time.
Additionally, it shows the most similar appearing face obtained from the similarity search network.}
\label{fig:test_screen}
\end{figure}

The straightforward approach would sequentially first detect all faces, then estimate their age, gender, facial expression, and facial similarity; show the result on the screen and start over with the detection.
However, the refresh rate on screen would be dictated by the sum of execution times of individual components.
On the other hand, users are less sensitive to a slow refresh rate of age estimates than the slow refresh rate of the display itself.
Therefore, the system has to prioritize the tasks differently while maximizing the performance and minimizing idle times.

Our system consists of a screen, a camera, and a computer, and it estimates the age, gender, and facial expression of all faces seen by the camera. 
In addition to these functions, the most similar-looking face from a database of celebrity faces is shown next to the detected face.
Apart from serving as an illustrative example of modern human-level machine learning for the general public, the system also highlights several common aspects in real-time machine learning systems.
The subtasks needed to achieve these recognition results represent a wide variety of tasks, including (a) face detection, (b) age estimation, (c) gender prediction, (d) facial expression prediction, and (e) image retrieval. Moreover, all these tasks should operate in unison, such that each task will receive enough resources from a limited pool.

Overall, we make the following contributions:
\begin{itemize}
\item We present a detailed system-level architecture for estimating several attributes from facial images.
\item We show the real-time performance of each component of the proposed architecture and its smooth functionality even on a moderate-resourced computing platform.
\item We release source code and trained models, with detailed instructions for deployment.
\end{itemize}

\begin{figure*}[t]
\centering
\includegraphics[width=0.80\textwidth]{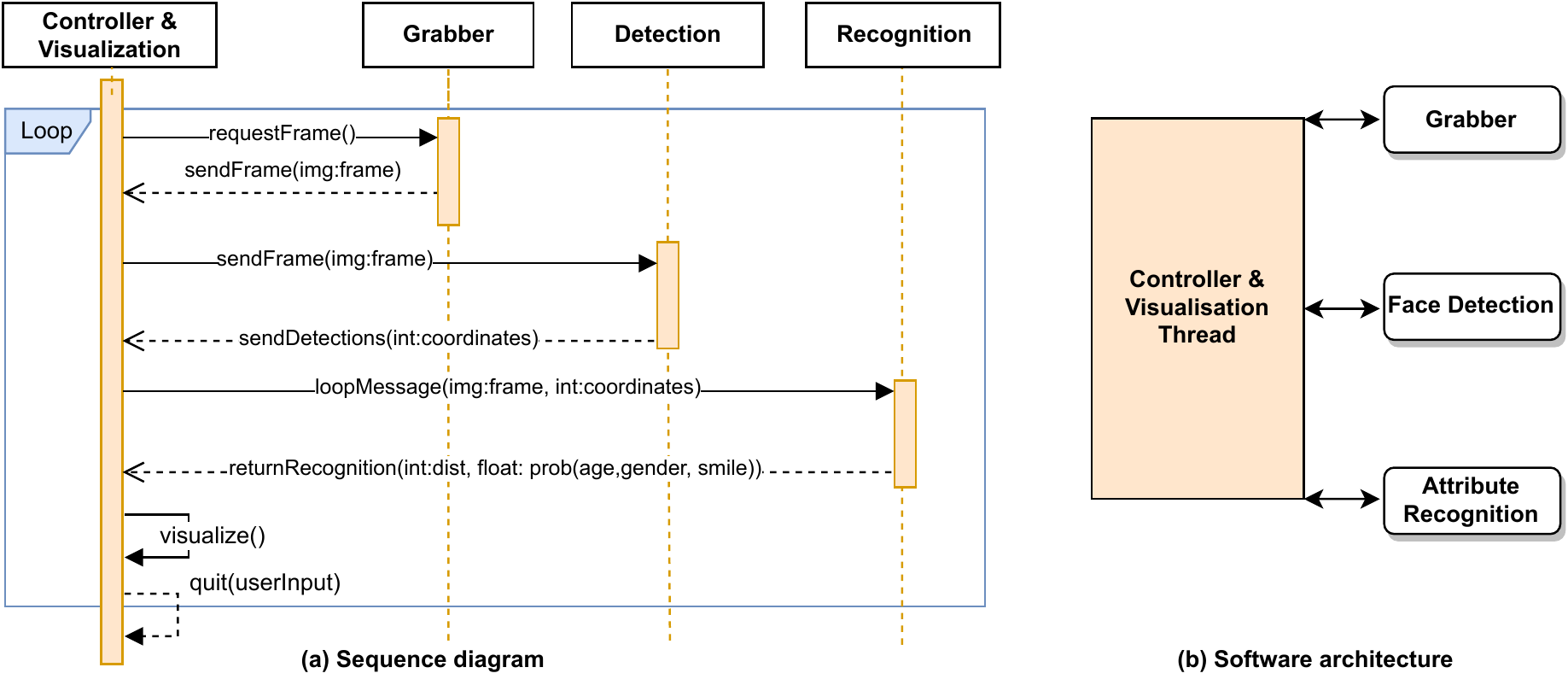}
\caption{Sequence diagram of the proposed real-time facial analysis system in (a) and software architecture of our system in (b).}
\label{fig:software_architecture}
\end{figure*}

The structure of the rest of the paper is as follows.
In Section~\ref{sec:system label functionality} we describe the system level multi-threaded architecture for real-time processing.
This is followed by a detailed description of individual components of the system in Section~\ref{sec:system components}.
Next, we report the experimental setups together with datasets and performance measurement metrics in Section~\ref{sec:experiments}.
We present experimental results of each recognition component in Section~\ref{sec:results and discussion} and finally, we discuss the benefits of demonstrating the potential of modern machine learning to both the general public and experts in the field.

\section{System Level Functionality}
\label{sec:system label functionality}

The challenge in real-time operation is that there are multiple components in the system, and each uses a different amount of execution time.
The system should be designed such that the operation appears smooth, which means that the most visible tasks should be given higher priority in scheduling.

The implementation is multi-threaded, as illustrated in Fig.~\ref{fig:software_architecture}.
Each thread operates asynchronously, with \textit{recognition threads} polling for new frames to process whenever they are idle.
The system is controlled by the \textit{controller \& visualization thread}, which receives new frames from the camera via the dedicated \textit{grabber thread}.
The controller thread also stores the frames in a buffer with each frame associated with flags, whether they have been processed by each of the threads.
Finally, it visualizes by showing the live video as well as overlay the most recent recognition results to the user in real-time.
The asynchronous threading structure also allows execution on dedicated platforms (\textit{e.g.}, detection running on the CPU and recognition on the GPU).
Also, it enables straightforward process prioritization by launching multiple recognition threads for the same task.

\subsection{Frame Capture}
The recognition process starts from the \textit{grabber thread}, which is connected to a camera.
The thread receives video frames from the camera for feeding them into a memory buffer located inside the controller thread.
At grab time, each frame is wrapped inside a class object, which holds the necessary metadata: a time-stamp and flags indicating whether each of the processing stages (face detection, attributes recognition, and similarity search) has been applied on the frame.

\subsection{Face Detection}

The first processing step in the pipeline is to find all faces in the input frame.
The detection is executed in a dedicated thread, which operates asynchronously, continuously requesting new non-processed frames from the controller thread.
The detection algorithm is discussed in detail in Section~\ref{sec:detection}.
Finally, the coordinates of the bounding boxes of all found faces are sent to the controller thread.
The controller thread stores the locations and matches each new face with all face objects from the previous frames using straightforward centroid tracking.
Tracking allows the system to temporally average the estimates (age, gender, and smile) for each face over a number of recent frames to improve the resulting accuracy.

\subsection{Facial Attributes Recognition}

The recognition thread is responsible for assessing the age, gender, facial expression, and facial similarity of each face crop found from the image.
Like the detection thread, the recognition thread also operates in an asynchronous mode, requesting new non-processed (but face-detected) frames from the controller thread.
When a new frame is received, the thread first aligns the face with a face template.
After alignment, we pass each aligned face to separate networks: age, gender, and expression recognizer or a multitask and a similarity search.

Typically, the networks executed on the face crops are slower than the detection network.
On the other hand, the amount of time grows linearly with the number of detected faces in the scene.
Therefore, in order for the system to appear fast and responsive, these tasks should run in the background and only refresh when each task finishes.
More specifically, we refresh the camera view and face detection in real-time but update the recognition results at less than the real-time rate.
Moreover, the recognition thread prioritizes the facial expression task over others because age, gender, and facial similarity can be assumed to be constant, while users expect a quick response to their expressions.

The system is implemented using the TensorFlow and OpenCV libraries.
The proposed facial analysis architecture can run on various hardware configurations, exploiting either CPU or GPU hardware.
As shown in Section~\ref{sec:results and discussion}, common desktop hardware reaches real-time speed both on CPU and GPU.
However, if the camera resolution, detector type, or input resolution are changed, then a GPU can be used instead.

\section{System Components}
\label{sec:system components}
\subsection{Face Detection}
\label{sec:detection}

Face detection is the first step for facial recognition systems, where the location of the face is extracted from the given image.
We design a neural network based face detector trained using benchmark face datasets.
The detectors are not initialized from scratch but fine-tuned from existing pre-trained weights.
We experimented with several models from two neural network based detection model categories: single-stage and two-stage detection networks.
The single-stage Single Shot Detector(SSD)~\cite{liu2016ssd}  requires only a single pass through the network with the image as the input and target bounding boxes with respective confidences as the outputs.
The two-stage Regions Convolutional Neural Network (RCNN)~\cite{ren2015faster} operates in two stages: a region proposal network proposes candidate object locations, followed by a classifier that classifies the proposals to target categories.

These two structures represent two widely used architectures, where the two-stage RCNN is traditionally perceived as more accurate, especially with small targets.
On the other hand, the SSD type networks are simpler, reach faster execution time, and still achieve a reasonable accuracy when the targets are not exceptionally small.
However, recent improvements~\cite{FPN_paper, retinanet_paper} in single-stage detectors have brought single-stage and two-stage architectures closer to each other, both in terms of accuracy and execution speed.

SSD model together with feature extractor networks such as MobileNetV1~\cite{MobileNet} and MobileNetV2~\cite{MobileNet_V2} are popular for faster and light-weight object detection.
MobileNetV1 introduced a parameter $\alpha$ called width multiplier to build a smaller and computationally efficient model.
This width multiplier has the effect of reducing computational cost and the number of parameters quadratically by roughly $\alpha^2$ times.
MobileNetV2 introduced a mobile-friendly variant SSDLite that replaces regular convolutions with separable convolutions in the SSD prediction layers, reducing both parameter count and computational cost.

\subsection{Alignment}
\label{sec:align}

\begin{figure}[t]
    \centering
    \includegraphics[width=0.32\textwidth]{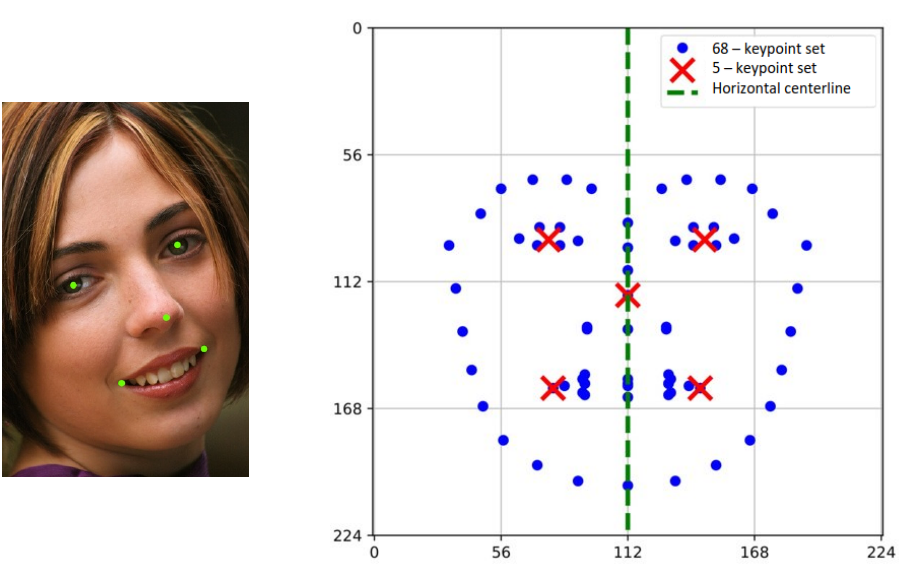}
    \caption{An example of five-point facial keypoint on a cropped face region (left) and keypoint template (right). Symmetric keypoints are in blue dots, and the 5 referenced true keypoints are highlighted with orange color.}
    \label{fig:keypoints}
\end{figure}

We align the faces in two stages.
The first stage locates a set of facial keypoints from the face crop: eyes, nose, and the corners of the mouth.
In the second stage, we find an affine mapping between these five keypoint locations and the corresponding template of five keypoints. This improves accuracy since the recognizers always see the eyes, mouth, and other facial elements in fixed locations, and require less effort in understanding the context where facial features are located.
This also enables the use of smaller networks, which compensate for the added computation due to the alignment procedure.

\textbf{\em Keypoint Detection}---%
The intention of aligning the faces to fixed coordinates is that this should improve the prediction accuracy.
To this aim, we first find the keypoints for each face detected by the detector.
We use five facial keypoints for normalizing the face location: eyes, nose, and the corners of the mouth, as illustrates in Fig.~\ref{fig:keypoints}.
Among the accurate and lightweight keypoint detection techniques, we consider regression forests of Kazemi \textit{et al.}~\cite{kazemi2014one} and a convolutional neural network, where both receive the face crop as input and output the predicted x-y-coordinates of the five keypoints.
We design the convolutional network according to the keypoint location branch of the O-Net~\cite{cnn_alignmnet_o_net}; consisting of four convolutional layers and two fully connected layers.
The facial keypoint detector is trained from scratch on AFLW dataset~\cite{AFLW}.

\textbf{\em Affine Mapping}---%
The detected keypoints are aligned to a set of template keypoints.
The template is obtained from the keypoints of a randomly selected sample face from the dataset.
However, we normalize the template such that the keypoints are horizontally symmetric with respect to the centerline of the face.
This is done in order to allow training set augmentation by adding horizontal flips of each training face.
More specifically, we manually marked symmetric pairs of keypoints and averaged their vertical coordinates and distances from the horizontal center location as illustrated in Fig.~\ref{fig:keypoints}.
Finally, the resulting set of coordinates was scaled to fit the network input size of $224\times 224$ pixels, leaving 10\% margin at the bottom edge and 20\% margin at the other edges.

\noindent \textbf{\em Face Alignment}---Instead of the simple approach of using the full affine transformation with least squares fit, we choose to use a more restricted \emph{similarity transformation} allowing only rotation, scale, and translation, but not shearing.
This is due to the possible distortion of the facial shape and the subsequent degradation of the estimation performances.

The similarity transformation $\boldsymbol{H}$ that maps 2D coordinate points $\ub\in\R^2 \mapsto \vb\in\R^2$ with translation $\tb=(t_x,t_y)^T$, scaling $s\in \R_+$ and rotation matrix ${\bf R}$ with rotation angle $\theta\in [-\pi, \pi]$ is given by
\begin{equation}
\vb=\boldsymbol{H}\x=
\begin{bmatrix}
    s\boldsymbol{R}       & \tb   \\
    \boldsymbol{0}^{T}      & 1   \\
\end{bmatrix}\ub =
\begin{bmatrix}
    s\cos\theta    & -s\sin\theta & t_{x}  \\
    s\sin\theta      & s\cos\theta   & t_{y}\\
    0      & 0   & 1\\
\end{bmatrix} \ub  \label{eqn.similarity}
\end{equation}
Estimation of the transformation parameters--- ${\bf R}$, ${\bf t}$ and $s$--- can be obtained from the vector cross product of point correspondences in homogeneous coordinates~\cite{hartley2003multiple}. Given x-y-coordinates of detected keypoints $\ub_i=(x_i,y_i,1)^{T}$ and corresponding template locations $\vb_i=(x'_i,y'_i,1)^{T}$ for $i = 1,2,\ldots, P$ (with at least $P = 2$ correspondences), the least squares solution for $\boldsymbol{H}$ can be obtained from the equation
\begin{equation}
\vb_i\times\boldsymbol{H}\ub_i=0.
\label{eqn.dlt}
\end{equation}
Substituting Eq. (\ref{eqn.similarity}) into Eq. (\ref{eqn.dlt}), the system is further simplified to ~\cite{KamPaa:2009}
\begin{equation}
\begin{bmatrix}
    -y_i   & -x_i & 0 & 1 \\
    x_i      & -y_i   & 1& 0\\
\end{bmatrix}
\begin{bmatrix}
    s\cos\theta  \\
    s\sin\theta \\
    t_x \\
    t_y \\
\end{bmatrix}=
\begin{bmatrix}
    -y'_i \\
    x'_i\\
\end{bmatrix},
\end{equation}
which can be solved by the singular value decomposition~\cite{hartley2003multiple}.
Finally, we construct the similarity matrix $\boldsymbol H$ by inserting the four solved scalar unknowns into it. 

\subsection{Age Estimation}
\label{sec:age}
Age estimation is commonly treated as a regression problem.
However, in our system, we treated this as a classification task as our system predicts ages among 101 classes.
The network is initialized using ImageNet~\cite{deng2009imagenet} pre-trained weights and fine-tuned in two stages: first with the large but noisy 500K IMDB-WIKI dataset~\cite{imdb_wiki} and then using the small but accurate CVPR2016 LAP challenge dataset~\cite{LAP2016}.

\subsection{Gender and Expression Recognition}
\label{sec:gender}
The gender recognition network is trained from scratch in two stages: first with the 500K IMDB-WIKI dataset and then fine-tuned with the CVPR2016 LAP challenge dataset, same as in the age recognition step.

In our system, we focused only on smile recognition, a binary classification task, detecting smile and non-smile.
The smile recognition network is initialized with ImageNet pre-trained weights and fine-tuned with Genki4k dataset~\cite{GENKI4K}.

\subsection{Facial Similarity Search}
\label{sec:similarity}
In addition to the age, gender, and facial expression, the fourth analysis task integrated into the system is the facial similarity search.
It is currently implemented for demonstrating \emph{celebrity search}, \emph{i.e.}, the program holds a database of celebrity faces and displays the one whose face has the most similar appearance to the person in front of the demo system.
Alternatively, this functionality could be altered to keep track of users using a dynamic database (persons are added to the database every time they are seen) instead of a fixed database (a static collection of celebrities).

The facial similarity search is implemented in two stages:
(1) the first stage computes a feature vector from the facial crop using a convolutional network,
and (2) the second stage performs the nearest neighbor search among the database of precomputed feature vectors from celebrity faces.
We use the FAISS implementation from Facebook~\cite{faiss}, since it is widely adopted, provides an interface in Python, and satisfies the real-time speed requirement even with large databases.

We adopt a person re-identification framework~\cite{luo2019bag} to find the most similar face from a collection of celebrity faces.
The backbone models are initialized with ImageNet pre-trained weights, and a global average pooling layer is appended to squeeze the spatial dimensions.
Several data augmentation policies are applied to make the model more robust, including random flipping, cropping, and random erasing~\cite{zhong2017random}.
At the early stage of the training, the learning rate starts from a relatively low value and increases gradually.
Additionally, the learning rate gets reduced to one-tenth once the performance on the validation split plateaus.

\subsection{Multitask Network}
\label{sec:multitask}

\begin{figure}[t]
    \centering  
     \includegraphics[width=0.35\textwidth]{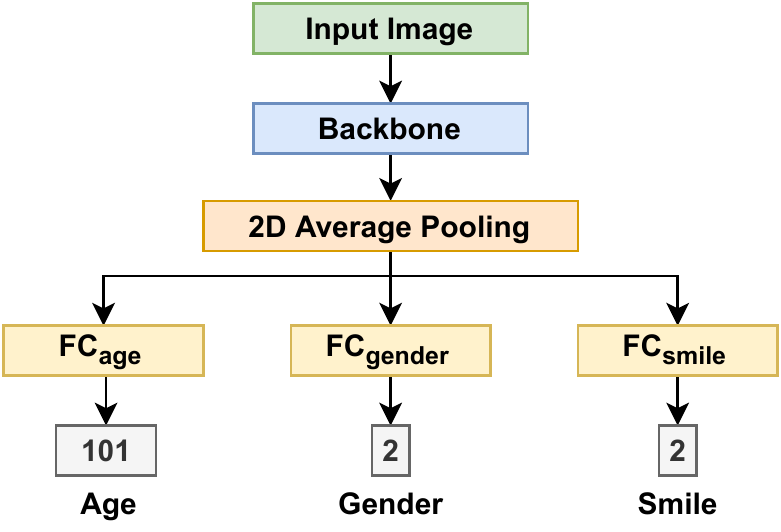}
    \caption{The architecture of our multitask classification network. The network is able to classify age, gender, and smile attributes for a given image.}
    \label{fig:multitask}
\end{figure}

We experimented with a single multitask network architecture shown in Fig.~\ref{fig:multitask} for age, gender, and smile predictions.
The multitask network utilizes a transfer learning approach; backbone network (ImageNet trained) weight is used to fine-tune on the age, gender, and smile training data.
The backbone network can be any neural network for the classification task.
The last layer of the network is removed, the global average pooling layer is added, and the output from the pooling layer is split into three branches.
The fully connected layers, each of dimension 512 is added and SoftMax of different dimensions is applied for task-specific output branches.

Our multitask network inference time is almost identical to individual classification networks;
hence this network is about three times faster than the individual networks for age, gender, and smiles recognition tasks, as reported in Table~\ref{tab: multitask results}.

\section{Experiments}
\label{sec:experiments}

\subsection{Datasets}

\noindent \textbf{\em AFLW}--- The Annotated Facial Landmarks in the Wild (AFLW)~\cite{AFLW} is a large-scale dataset of (25K) face images collected from Flickr. It has 21 landmarks annotations per face. We use this dataset to train 5 -- keypoint set detection.

\noindent \textbf{\em CelebA}--- The CelebFaces Attributes (CelebA)~\cite{celebA} is a large-scale face attributes dataset containing more than 200K celebrity images from 10,177 identities. It has 5 landmark locations and 40 binary attribute labels per facial image.

\noindent \textbf{\em ChaLearn LAP}--- 
We use the CVPR2016 competition variant, consisting of 7,591 facial images with human-annotated apparent ages and standard deviations taken in non–controlled environments with diverse backgrounds.

\noindent \textbf{\em Genki-4k}--- The MPLab Genki-4k~\cite{GENKI4K} contains 4,000 images with two class expressions (smile or non-smile) labeled by human and head-pose labels of the faces determined by automatic face detector.

\subsection{Evaluation Metrics}
\noindent \textbf{\em AP}---%
The Average Precision (AP) metric computes the average precision overall detection thresholds.
The sensitivity of the detector can be adjusted using a detection threshold set by default at 0.5.
As the sensitivity of detection may be adjusted at the inference process, we also average the class-wise AP's over all classes to produce the mean AP (mAP).

\noindent \textbf{\em MAE}---%
The Mean Absolute Error (MAE) metric computes the average error overall prediction.
We used MAE to measure the error (in years) at the age prediction stage.

\noindent \textbf{\em Accuracy}---%
Accuracy is the fraction of correctly classified instances among the total number of instances.

\noindent \textbf{\em CMC rank-k accuracy}---%
Given a query sample, the accuracy is set to 1 if the top-k gallery samples contain samples that have the same identity as the query sample, and 0 otherwise.
The CMC rank-k accuracy is obtained by averaging the results of all query samples.
\section{Results and Discussion}
\label{sec:results and discussion}

\begin{table}[t]
\begin{threeparttable}
\centering
\caption{Comparison of different detection models for face detection with different input sizes.}
\label{tab:detectionmodels}
\begin{tabular}{cccccc}
\toprule
\multirow{2}{*}{Resolution} &{AP} &{AP} & {FPS} &{FPS}&{FPS} \\ 
&0.5:0.95 &@0.5 &TF-CPU &TF-GPU &OpenCV  \\ \hline
\midrule
\multicolumn{6}{@{}l}{\textbf{Faster RCNN ResNet101}} \\
300x300 & \textbf{0.747} & \textbf{0.945} & 1.84 & 7.62 & 1.09 \\
240x180 & 0.707 & 0.914 & 1.93 & 8.18 & 1.20 \\
200x200 & 0.693 & 0.907 & \textbf{1.98} & \textbf{8.30} & \textbf{1.21} \\
\midrule
\multicolumn{6}{@{}l}{\textbf{SSD MobileNetV1 $\alpha = 1$}} \\
300x300 & \textbf{0.744} & \textbf{0.945} & 32.52 & 87.29 & 39.36 \\
240x180 & 0.684 & 0.868 & \textbf{51.60} & 105.46 & 70.09 \\
200x200 & 0.683 & 0.839 & 49.96 & \textbf{107.54} & \textbf{71.42} \\ 
\midrule
\multicolumn{6}{@{}l}{\textbf{SSD MobileNetV1 $\alpha = 0.25$}} \\
300x300 & \textbf{0.695}  & \textbf{0.909}  & 60.50 & 148.77  & 140.95  \\
240x180 & 0.647  & 0.895  & \textbf{81.29}  & 156.62  & 239.59 \\
200x200 & 0.650  & 0.887  & 78.21  & \textbf{158.20}  & \textbf{249.72} \\ 
\midrule
\multicolumn{6}{@{}l}{\textbf{SSDLITE MobileNetV2 $\alpha = 1$}} \\
300x300 & \textbf{0.764}  & \textbf{0.952}  & 28.46  & 79.94  & 36.47 \\
240x180 & 0.728  & 0.936  & \textbf{41.18}  & 106.98  & 63.29 \\
200x200 & 0.730  & 0.934  &40.50  & \textbf{109.47}  & \textbf{64.79} \\ 
\midrule
\multicolumn{6}{@{}l}{\textbf{SSDLITE MobileNetV2 $\alpha = 0.25$}} \\
300x300 & \textbf{0.733} & \textbf{0.936}  & 43.09  & 118.29  & 70.58  \\
240x180 & 0.704 & 0.925  & 60.43  & 129.25  & 125.55  \\
200x200 & 0.679 & 0.913  & \textbf{64.01}  & \textbf{131.77}  & \textbf{131.22 } \\ 
\bottomrule
\end{tabular}
\end{threeparttable}
\end{table}

\subsection{Face Detection} 
For face detection, we use faster RCNN, with ResNet101 backbone and variants of SSD, with MobileNet backbones with three different input sizes. The network inference speed is tested on Tensorflow CPU and GPU, and OpenCV environments, illustrated in Table~\ref{tab:detectionmodels}.  The faster RCNN ResNet101 network in all experiments gives slightly higher AP than SSD models.
However, the computation complexity of RCNN models is high, i.e., lower FPS compared to one-stage networks.
Experiments show that with $\alpha = 0.25$, detection performance is about 3\% less accurate while increasing inference speed about 1.5 times.

For all experimented networks, optimal detection performance is obtained by larger input size (\textit{i.e.}, $ 300 \times 300$), while best FPS is obtained with smaller input size (\textit{i.e.}, $ 200 \times 200$).
With a smaller square input size, using OpenCV at inference always guarantees the best inference speed.
If detection accuracy is not the top priority, using a small value of the $ \alpha$, smaller input size, and lighter model is suitable for faster and memory-efficient detection.

We measured the performance of two keypoint detection methods on the AFLW dataset.
During the training,  keypoint detection models were optimized based on the detected facial area with ground-truth keypoint labels.

Experiments on O-Net~\cite{cnn_alignmnet_o_net} based CNN alignment gives better alignment accuracy and inference speed as shown in Table~\ref{tab:table_keypoints}.

\begin{table}[t]
\begin{minipage}[b]{0.2\textwidth}
    \centering
    \begin{threeparttable}
    \caption{Keypoint detection performance.}
    \label{tab:table_keypoints}
    \begin{tabular}{lcc}
    \toprule
     & Error  & FPS \\
     & rate(\%) & CPU\\ \hline
    \midrule
    Dlib & 2.89 & 17.49 \\
    CNN & \textbf{1.04}  & \textbf{18.25} \\
    \bottomrule
    \end{tabular}
    \end{threeparttable}
\end{minipage}%
\begin{minipage}[b]{0.3\textwidth}
    \centering
    \begin{threeparttable}
    \caption{Performance in facial similarity using three  backbones.}
    \label{tab:similarity_search}
    \begin{tabular}{lccc}
    \toprule
    Network & mAP & rank-1 & rank-5 \\ \hline
    \midrule
    MobileNet & 0.782 & 0.940 & 0.970 \\
    VGG16 & 0.813 & 0.952 & 0.973 \\
    ResNet50 & \textbf{0.822} & \textbf{0.953} & \textbf{0.973} \\
    \bottomrule
    \end{tabular}
    \end{threeparttable}
\end{minipage}
\vspace{-\baselineskip}
\end{table}

\subsection{Facial Similarity}

Our facial similarity system aims at finding the most similar face, and the rank-1 accuracy is a preferable evaluation metric.
Table~\ref{tab:similarity_search} shows the mAP, rank-1 accuracy, and rank-5 accuracy of facial similarity models trained with categorical cross-entropy loss on the aligned images from the CelebA dataset.
The rank-1 accuracy of MobileNet reaches 94.0\% which is slightly inferior to VGG16 and ResNet50, while using MobileNet is computationally lightweight.

\subsection{Age, Gender and Expression Recognition}
Table~\ref{tab:accuracies_of_different_networks} shows the accuracies of the different tasks included in our system.
The speed test of each task on two different environments indicates that in the same environment, there is no significant difference between the network in terms of inference speed.
However, the multitask network appears better considering the total inference time for three tasks.

\begin{table}[t]
\centering
\begin{threeparttable}
\caption{Accuracies and inference speed at different stages in our system. The depth multiplier $\alpha = 1.0$ is used in all MobileNetV1 networks.}
\begin{tabular}{lcccc}
\toprule
\begin{tabular}[c]{@{}c@{}}Stage\\Network\end{tabular} & Accuracy  & \begin{tabular}[c]{@{}c@{}}FPS\\CPU\end{tabular} & \begin{tabular}[c]{@{}c@{}}FPS\\1050 TI\end{tabular} & \begin{tabular}[c]{@{}c@{}}FPS\\1080 TI\end{tabular} \\
\hline
\midrule
\begin{tabular}[c]{@{}c@{}}Age\\MobileNetV1\end{tabular}  & 4.9 MAE & 31.61 & 148.90 & 147.44 \\
\midrule
\begin{tabular}[c]{@{}c@{}}Gender\\MobileNetV1\end{tabular}  & 88.3\% & 31.48 & 150.45 & 149.75 \\
\midrule
\begin{tabular}[c]{@{}c@{}}Smile\\MobileNetV1\end{tabular}  & 87.2\% & 31.46 & 148.84 & 148.78\\
\midrule
\begin{tabular}[c]{@{}c@{}}Multitask\\MobileNetV1\end{tabular}  &\begin{tabular}[c]{@{}c@{}}5.67 MAE\\ 84.2\% Gender \\ 83.6\% Smile \end{tabular} & 29.80 & 147.06 & 147.20 \\
\midrule
\begin{tabular}[c]{@{}c@{}}Multitask\\EfficientNetB0\end{tabular}  &\begin{tabular}[c]{@{}c@{}} 5.35 MAE\\  87.5\% Gender \\ 86.0\% Smile \end{tabular} & 25.61 & 143.35 & 144.24 \\
\bottomrule
\end{tabular}
\label{tab:accuracies_of_different_networks}
\end{threeparttable}
\end{table}

The experimental results on our multitask network for age estimation, gender, and smile recognition with different backbone networks are reported in Table~\ref{tab: multitask results}. The best performing multitask network gives 5.35 years age MAE which is slightly higher than the best performing individual age network.
Also, gender and smile accuracies obtained from the best performing multitask network are slightly less accurate than the individual networks.
EfficientNet~\cite{efficientNets_2019} networks give better age MAE and recognition accuracies, but they are computationally expensive.
As our computational budget is limited, and we cannot use a combination of many networks.

We can set up a system with a combination of a lighter detection model and a faster multitask network if the network accuracies are not the top priority.
This way, real-time inference speed can be achieved on the CPU while slightly compromising each task's accuracy.

\begin{table}[t]
\centering
\begin{threeparttable}
\caption{Performance comparison of the multitasking network with different backbone networks.}
\label{tab: multitask results}
\begin{tabular}{lcccc}
\toprule
\multirow{2}{*}{Network} &Age & Gender & Smile & FPS \\ 
& MAE & Acc(\%) & Acc(\%) & CPU \\\hline
\midrule
VGG16           & 7.20  & 84.0  & 84.1  &  27.75 \\ 
ResNet50        & 6.42  & 82.1  & 81.2  &  27.06 \\
ResNet18        & 6.02  & 82.4  & 82.8  &  29.63 \\
MobileNetV1    & 5.67  & 84.2  & 83.6  &  \textbf{29.80} \\
EfficientNet B0 & 5.35  & 87.5  & 86.0  &  25.61 \\
EfficientNet B1 & 5.07  & 87.8  & 86.8  &  22.72 \\
EfficientNet B7 & \textbf{4.37}  & \textbf{89.5}  & \textbf{87.3}  &  14.63 \\
\bottomrule
\end{tabular}
\end{threeparttable}
\end{table}

\section{Conclusion}
\label{sec:concl}
We present a system-level design of a human facial analysis system with a multi-threaded architecture to reach real-time operation on resource-limited devices. We describe individual components of our system, integrating several standard machine learning components with an extensive set of experiments on each task. Users can switch specific task networks from the list of available options on the fly.
The demo system has been presented several times in public locations. It has shown its value in illustrating the potential of modern machine learning in an easy-to-approach use case working on many levels.
Moreover, this system can be used in the surveillance system by adding an alarm function that triggers once the detected face is matched with the suspect's faces on the query dataset.
Additionally, the system can be used as a reference and baseline for related applications.

\bibliographystyle{IEEEtran}
\bibliography{reference}

\end{document}